# Composite Monte Carlo Decision Making under High Uncertainty of Novel Coronavirus Epidemic Using Hybridized Deep Learning and Fuzzy Rule Induction

Simon James Fong[1,2*], Gloria Li[2], Nilanjan Dey[3†], Rubén González Crespo[4], Enrique Herrera-Viedma[5]

[1]*Department of Computer and Information Science, University of Macau, Macau SAR, China*
[2]*DACC Laboratory, Zhuhai Institutes of Advanced Technology of the Chinese Academy of Sciences, China*
[3]*Department of Information Technology, Techno India College of Technology, India*
[4]*Universidad Internacional de La Rioja, Logroño, Spain*
[5]*University of Granada, Spain*

**Abstract**

In the advent of the novel coronavirus epidemic since December 2019, governments and authorities have been struggling to make critical decisions under high uncertainty at their best efforts. Composite Monte-Carlo (CMC) simulation is a forecasting method which extrapolates available data which are broken down from multiple correlated/casual micro-data sources into many possible future outcomes by drawing random samples from some probability distributions. For instance, the overall trend and propagation of the infested cases in China are influenced by the temporal-spatial data of the nearby cities around the Wuhan city (where the virus is originated from), in terms of the population density, travel mobility, medical resources such as hospital beds and the timeliness of quarantine control in each city etc. Hence a CMC is reliable only up to the closeness of the underlying statistical distribution of a CMC, that is supposed to represent the behaviour of the future events, and the correctness of the composite data relationships. In this paper, a case study of using CMC that is enhanced by deep learning network and fuzzy rule induction for gaining better stochastic insights about the epidemic development is experimented. Instead of applying simplistic and uniform assumptions for a MC which is a common practice, a deep learning-based CMC is used in conjunction of fuzzy rule induction techniques. As a result, decision makers are benefited from a better fitted MC outputs complemented by min-max rules that foretell about the extreme ranges of future possibilities with respect to the epidemic.

---

* Corresponding co-author. *E-mail address:* ccfong@umac.mo,
* Corresponding co-author. *E-mail address:* nilanjan.dey@tict.edu.in,

# 1. Introduction

On top of devastating health effects, epidemic impacted hugely on world economy. In the Ebola outbreak between 2014-2016 where more than 28,000 and cases were suspected and 10,000 deaths in West Africa [1], $2.2 billion was lost [2]. On the other hand, SARS took over 648 lives from China including Hong Kong and 700 lives worldwide between 2002 and 2003 [3]. Its losses on global economy are up to a huge $100 billion, 1% and 0.5% dips of GDPs in Chinese and Asian domestic markets respectively [3]. Although the current coronavirus (codename: NCP or COVID-19) epidemic is not over yet, its economy impact is anticipated by economists from IHS Markit to be far worse than that of SARS outbreak in 2003 [4]. The impact is so profound that will lead to factories shut down, enterprises bankruptcy especially those in tourism, retail and F&B industries, and suspensions or withdrawals in long-term investment, if the outbreak cannot be contained in time. Since the first case in December 2019, the suspected cases and deaths around the world skyrocketed to over 76395 confirmed cases and 2348 deaths, mostly in China, by the time of writing this article.

An early intervention measure in public health to thwart the outbreak of COVID-19 is absolutely imperative. According to a latest mathematical model that was reported in research article by The Lancet [5], the growth of the epidemic spreading rate will ease down if the transmission rate of the new contagious disease can be lowered by 0.25. Knowing that the early ending the virus epidemic or even the reduction in the transmission rate between human to human, all governments especially China where Wuhan is the epicenter are taking up all the necessary preventive measures and all the national efforts to halt the spread. How much input is really necessary? Many decision makers take references from SARS which is by far the most similar virus to COVID-19. However, it is difficult as the characteristics of the virus are not fully known, it details and about how it spreads are gradually unfolded from day to day. Given the limited information on hand about the new virus, and the ever evolving of the epidemic situations both geographically and temporally, it boils down to grand data analytics challenge this analysis question: how much resources shall be enough to slow down the transmission? This is a composite problem that requires cooperation from multi-prong measures such as medical provision, suspension of schools, factories and office, minimizing human gathering, limiting travel, strict city surveillance and enforced quarantines and isolations in large scales. There is no easy single equation that could tell the amount of resources in terms of monetary values, manpower and other intangible usage of infrastructure; at the same time there exist too many uncertain variables from both societal factors and the new development of the virus itself. For example, the effective incubation period of the new virus was found to be longer than a week, only some time later after the outbreak. Time is an essence in stopping the epidemic so to reduce its damages as soon as possible. However, uncertainties are the largest obstacle to obtain an accurate model for forecasting the future behaviours of the epidemic should intervention apply. In general, there is a choice of using deterministic or stochastic modelling for data scientists; the former technique based solely on past events which are already known for sure, e.g. if we know the height and weight of a person, we know his body mass index. Should any updates on the two dependent variables, the BMI will be changed to a new value which remains the same for sure no matter how many times the calculation is repeated. The latter is called probabilistic or stochastic model - instead of generating a single and absolute result, a stochastic model outputs a collection of possible outcomes which may happen under some probabilities and conditions.

*1.1. Background*

Deterministic model is useful when the conditions of the experiment are assumed rigid. It is useful to obtain direct forecasting result from a relatively simple and stable situation in which its variables are unlikely to deviate in the future. Otherwise, for a non-deterministic model, which is sometimes referred as probabilistic or stochastic, the conditions of a future situation under which the experiment will be observed, are simulated to some probabilistic behaviour of the future observable outcome. For an example of epidemic, we want to determine how many lives could be saved from people who are infected by a new virus as a composite result of multi-prong efforts that are put into the medical resources, logistics, infrastructure, spread prevention, and others; at the same time, other contributing factors also matter, such as the percentage of high-risk patients who are residing in that particular city, the population and its mobility, as well as the severity and efficacy of the virus itself and its vaccine respectively. Real-time tools like CDC data reporting and national big data centers are available with which any latest case that occurs can be recorded. However, behind all these records are sequences of factors associated with high uncertainty. For example, the disease transmission rate depends on uncertain variables ranging from macro-scale of weather and economy of the city in a particular season, to the individual's personal hygiene and the social interaction of commuters as a whole. They are dynamic in nature that change quickly from time to time, person to person, culture to culture and place to place. The phenomena can hardly converge to a deterministic model. Rather, a probabilistic model can capture more accurately the behaviours of the phenomena. So for epidemic forecast, a deterministic model such as trending is often used to the physical considerations to predict an almost accurate outcome, whereas in a non-deterministic model we use those considerations to predict more of a probable outcome that is probability distribution oriented.

In order to capture and model such dynamic epidemic recovery behaviours, stochastic methods ingest a collection of input variables that have complex dependencies on multiple risk factors. The epidemic recovery can be viewed in abstract as a bipolar force between the number of populations who has contracted the disease and the number of patients who are cured from the disease. Each group of the newly infested and eventually cured (or unfortunately deceased) individuals are depending on complex societal and physiological factors as well as preventive measures and contagious control. Each of these factors have their underlying and dependent factors carrying uncertain levels of risks. A popular probabilistic approach for modeling the complex conditions is known as Monte Carlo (MC) simulation which provides a means of estimating the outcome of complex functions by simulating multiple random paths of the underlying risk factors. Rather than deterministic analytic computation, MC uses random number generation to generate random samples of input trials to explore the behaviour of a complex epidemic situation for decision support. MC is particularly suitable for modeling epidemic especially new and unknown disease like COVID-19 because the data about the epidemic collected on hand in the early stage are bound to change. In MC, data distributions are entered as input, since precise values are either unknown or uncertain. Output of MC is also in a form of distribution specifying a range of possible values (or outcome) each of which has its own probability at which it may occur. Compared to deterministic approach where precise numbers are loaded as input and precise number is computed to be output, MC simulates a broad spectrum of possible outcomes for subsequent expert evaluation in a decision-making process.

*1.2. Monte Carlo Simulation for Epidemics*

Recently as epidemic is drawing global concern and costing hugely on public health and world economy, the use of MC in epidemic modeling forecast has become popular. It offers decision makers an extra dimension of probability information so called risk factors for analyzing the possibilities and their associated risk as a whole. Decades ago, there has been a growing research interest in using MC for quantitatively modelling epidemic behaviours. Since 1957, Bailey et al was among the pioneers in formulating mathematical theory of epidemics. Subsequently in millennium, Andersson and Britton [7] adopted MC simulation techniques to study the behaviour of stochastic epidemic models, observing their statistical characteristics. In 2003, House et al, attempted to estimate how big the final size of an epidemic is likely to be, by using MC to simulate the course of a stochastic epidemic. As a result, the probability mass function of the final number of infections is calculated by drawing random samples over small homogeneous and large heterogeneous populations. Yashima and Sasaki in 2013 extended the MC epidemic model from over a population to a particular commute network model, for studying the epidemic spread of an infectious disease within a metropolitan area - Tokoyo train station. MC is used to simulate the spread of infectious disease by considering the commuters flow dynamics, the population sizes and other factors, the proceeding size of the epidemic and the timing of the epidemic peak. It is claimed that the MC model is able to serve as a pre-warning system forecasting the incoming spread of infection prior to its actual arrival. Narrowing from the MC model which can capture the temporal-spatial dynamics of the epidemic spread, a more specific MC model is constructed by Fitzgerald et al [10] in 2017 for simulating queuing behaviour of an emergency department. The model incorporates queuing theory and buffer occupancy which mimic the demand and nursing resource in the emergency department respectively. It was found that adding a separate fast track helps relieving the burden on handling of patient and cutting down the overall median wait times during an emergency virus outbreak and the operation hours are at peak. Mielczarek and Zabawa [11] adopted a similar MC model to investigate how erratic the population is, hence the changes in the number of infested patients affect the fluctuations in emergency medical services, assuming there are epidemiological changes such as call-for-services, urgent admission to hospital and ICU usages. Based on some empirical data obtained from EMS center at Lower Silesia Region in Poland, the EMS events and changes in demographic information are simulated as random variables. Due to the randomness of the changes (in population sizes as people migrate out, and infested cases increase) in both demand and supply of an EMS, the less-structured model cannot be easily examined by deterministic analytic means. However, MC model allows decision makers to predict by studying the probabilities of possible outcomes on how the changes impact the effectiveness of the Polish EMS system. There are similar works which tap on the stochastic nature of MC model for finding the most effective escape route during emergency evacuation [12] and modelling emergency responses [13].

### 1.3. Motivation and Contributions

Overall, the above-mentioned related works have several features in common: their studies are centered on using a probabilistic approach to model complex real-life phenomena, where a deterministic model may fall short in precisely finding the right parameters to cater for every detail. The MC model is parsimonious that means the model can achieve a satisfactory level of explanation or insights by requiring as few predictor variables as possible. The model which uses minimum predictor variables and offers good explanation is selected by some goodness of fit as BIC model selection criterion. The input or predictor variables are often dynamic in nature whose values change over some spatial-temporal distribution. Finally, the situation in question, which is simulated by MC model, is not only complex but a prior in nature. Just like the new COVID-19 pandemic, nobody can tell when or whether it will end in the near future, as it depends on too many dynamic variables. While the challenges of establishing an effective MC model is acknowledged for modelling a completely new epidemic behaviour, the model reported in [13] inspires us to design the MC model by decomposing it into several sub-problems. Therefore, we proposed a new MC model called composite MC or CMC in short which accepts predictor variables from multi-prong data sources that have either correlations or some kind of dependencies from one another. The challenge here is to ensure that the input variables though they may come from random distribution, their underlying inference patterns must contribute to the final outcome in question.

Considering multi-prong data sources widen the spectrum of possibly related input data, thereby enhancing the performance of Monte Carlo simulation. However, naive MC by default does not have any function to decide on the importance of input variables. It is known that what matters for the underlying inference engine of MC is the historical data distribution which tells none or little information about the input variables prior to the running of MC simulation. To this end, we propose a pre-processor, in the form of optimized neural network namely BFGS-Polynomial Neural Network is used at the front of MC simulator. BFGS-PNN serves as both a filter for selecting important variables and forecaster which generates future time-series as parts of the input variables to the MC model. Traditionally all the input variables to the MC are distributions that are drawn from the past data which are usually random, uniform or some customized distribution of sophisticated shape. In our proposed model here, a hybrid input that is composed of both deterministic type and non-deterministic type of variables. Deterministic variables come from the forecasted time-series which are the outputs of the BFGS-PNN. Non-deterministic variables are the usual random samples that are drawn from data distributions. In the case of COVID-19, the future forecasts of the time-series are the predictions of the number of confirmed infection cases and the number of cured cases. Observing from the historical records, nevertheless, these two variables display very erratic trends, one of them contains extreme outliers. They are difficult to be closely modelled by any probability density function; intuitively imposing any standard data distribution will not be helpful to delivering accurate outcomes from the MC model. Therefore in our proposal, it is needed to use a polynomial style of self-evolving neural network that was found to be one of the most suited machine learning algorithm in our prior work [14], to render a most likely future curve that is unique to that particular data variable.

The composition of the multiple data sources is of those relevant to the development (rise-and-decline) of the COVID-19 epidemic. Specifically, a case of how much daily monetary budget that is required to struggle against the infection spread is to be modelled by MC. The data sources of these factors are publicly available from the Chinese government websites. More details follow in Section 2 below. The rationale behind using a composite model is that what appears to be an important figure, e.g. the number of suspected cases are directly and indirectly related to a number of sub-problems which of each carries different levels of uncertainty: how a person gets infested, depends on 1) the intensity of travel (within a community, suburb, inter-city, or oversea) 2) preventive measures 3) trail tracking of the suspected and quarantining them 4) medical resources (isolation beds) available, and 5) eventual cured or dead. Some of these data sources are in opposing influences to one another. For example, the tracking and quarantine measures gets tighten up, the number of infested drops, and vice-versa. In theory, more relevant data are available, the better the performance and more accurate outcomes of the MC can provide. MC plays an important role here as the simulation is founded on probabilistic basis, the situation and its factors are nothing but uncertainty. Given the available data is scare as the epidemic is new, any deterministic model is prone to high-errors under such high uncertainty about the future.

The contribution of this work has been twofold. Firstly, a composite MC model, called CMCM is proposed which advocates using non-deterministic data distributions along with future predictions from a deterministic model. The deterministic model in use should be one that is selected from a collection of machine learning models that is capable to minimize the prediction error with its model parameters appropriately optimized. The advantage of using both fits into the MC model is the flexibility that embraces some input variables which are solely comprised of historical data, e.g. trends of people infested. And those that underlying elements which contribute to the high uncertainty, e.g. the chances of people gather, are best represented in probabilistic distribution as non-deterministic variables to the MC model. By this approach, a better-quality MC model can be established, the outcomes from the MC model become more trustworthy. Secondly, the sensitivity chart obtained from the MC simulation is used as corrective feedback to rules that are generated from a fuzzy rule induction (FRI) model. It is known that FRI outputs decision rules with probabilities/certainty for each individual rule. A rule consists of a series of testing nodes without any priority weights. By referencing the feedbacks from the sensitivity chart, decision makers can relate the priority of the variables which are the tests along each sequence of decision rules. Combining the twofold advantages, even under the conditions of high uncertainty, decision makers are benefited with a better-quality MC model which embraces considerations of composite input variables, and fuzzy decision rules with tests ranked in priority. This tool offers a comprehensive decision support at its best effort under high uncertainty.

The remaining paper is structured as follow. Section 2 describes the proposed methodology called GROOMS+CMCM, followed by introduction of two key soft computing algorithms - BFGS-PNN and FRI which is adopted for forecasting some particular future trends as inputs to the MC model and generating fuzzy decision rules respectively. Section 3 presents some preliminary results from the proposed model followed by discussion. Section 4 concludes this paper.

## 2. A Novel Methodology

MC has been applied for estimating epidemic characteristics by researchers over the year, because the nature of epidemic itself and its influences are full of uncertainty. An application that is relatively less looked at but important is the direct cost of fighting the virus. The direct cost is critical to keep the virus at bay when it is still early before becoming a pandemic. But often it is hard to estimate during the early days because of too many unknown factors. Jiang et al [15] have modelled the shape of a typical epidemic concluding that the curve is almost exponential; it took only less than a week from the first case growing to its peak. If appropriate and urgent preventive measure was applied early to have it stopped in time, the virus would probably not escalate into an epidemic then pandemic. Ironically, during the first few days (some call it the golden critical hours), most of the time within this critical window was spent on observation, study, even debating for funding allocation and strategies to apply. If a effective simulation tool such as the one that is proposed here, decision makers can be better informed the costs involved and the corresponding uncertainty and risks. Therefore, the methodology would have to be designed in mind that the available data is limited, each functional component of the methodology in the form of soft computing model should be made as accurate as possible. Being able to work with limited data, flexible in simulating input variables (hybrid deterministic and its counterpart), and informative outcomes coupled with fuzzy rules and risks, would be useful for experts making sound decision at the critical time. Our novel methodology is based on Group of Optimized and Multisource Selection, (GROOMS) methodology [14] which is made for choosing a machine learning method which has the highest level of accuracy. GROOMS as a standalone optimizing process is in aid of assuring the deterministic model that is to be used as input variable for the subsequent MC simulation to have the most accurate data source input. By default, MC model at its naive form accepts only input variable from a limited range of standard data distributions (Uniform, Normal, Bernoulli, Pareto, etc.); best fitting curve technique is applied should the historical data shape falls out of the common data distribution types. However, this limitation is expanded in our composite MC model, so-called CMCM in such a way that all relevant data sources are embraced, both direct and indirect types. An enhanced version of neural network is used to firstly capture the non-linearity (often with high irregularity and lack of apparent trends and seasonality) of the historical data. Out of the full spectrum of data sources, direct and indirect, the selected data sources through feature selection by correlation, that are filtered by the neural network, whose data distributions would be taken as input variables to the MC model. The combined methodology, GROOMS+CMCM is shown in Figure 1.

Fig. 1. GROOMS+CMCM methodology

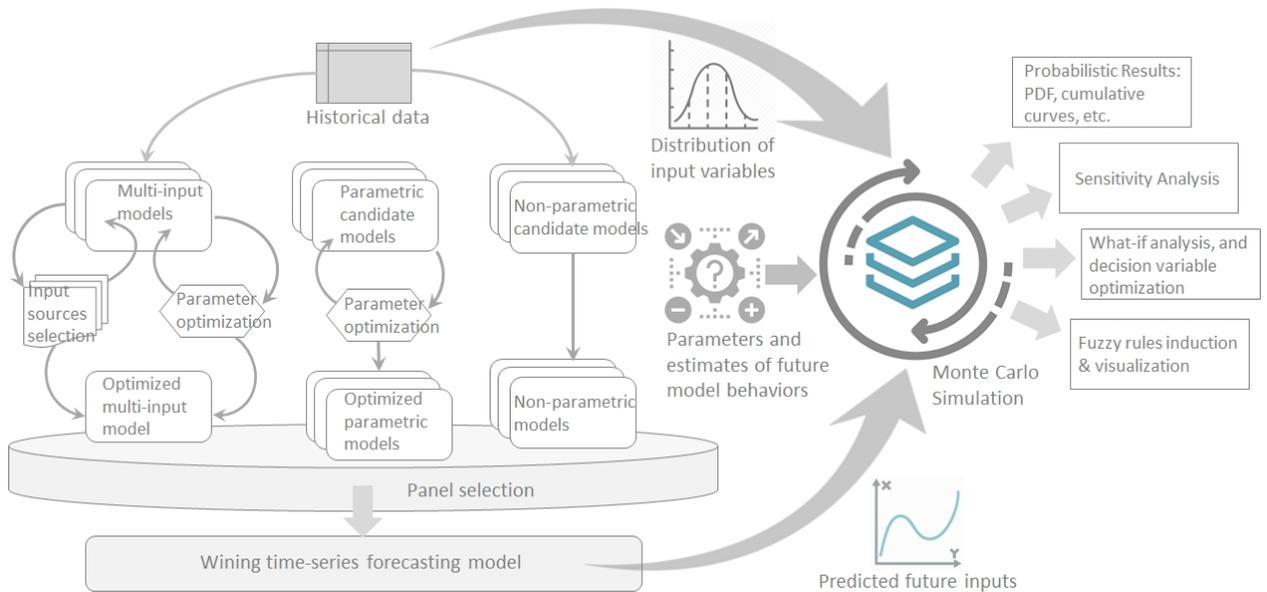

### 2.1. Broyden–Fletcher–Goldfarb–Shanno Optimized Polynomial Neural Network (BFGS-PNN)

According to the methodology, a machine learning algorithm candidate called BGFS-PNN which is basically PNN as selected as the winning algorithm in [14] enhanced with further parameter optimization function. The given time-series data fluctuated more than the same that were collected earlier. As a data pre-processor, BFGS-PNN has two functions. Firstly, for the non-deterministic data, using a *ClassifierBasedFilter* function in a wrapper approach, salient features could be found in feature selection. The selected salient features are those very relevant to the forecast target in the MC. In this case, it is composite MC model or CMCM as the simulation engine intakes multiple data sources from types of deterministic and non-deterministic. The second function is to forecast a future time-series as a type of deterministic input variable for the CMCM. The formulation of BFGS-PNN is shown as follow. The naïve version of PNN is identical to the one reported in [14]. BFGS-PNN uses BFGS (Broyden–Fletcher–Goldfarb–Shanno) algorithm to optimize the parameters and network structure size in an iterative manner using hill-climbing technique. BFGS theory is basically for solving non-linear optimization problem iteratively by finding a stationary equilibrium through Quasi-Newton method [16] and secant method [17]. Let PNN [18] take the form of Kolmogorov-Gabor polynomial as a functional series in Eqn. (1).

$$y = \varepsilon_0 + \sum_{t=1} \varepsilon_{t=1} x_{t=1} + \sum_{t=1}\sum_{t=2} \varepsilon_{t=1,t=2} x_{t=1} x_{t=2} + \sum_{t=1}\sum_{t=2}\sum_{t=3} \varepsilon_{t=1,t=2,t=3} x_{t=1} x_{t=2} x_{t=3} + \cdots \quad (1)$$

The polynomial is capable to take form of any function which is generalized as $y=f(\bar{x})$. The induction process is a matter of finding all the values for the coefficient vector $\bar{\varepsilon}$. As the process iterates, the variables from $\bar{x}$ arrive in sequence fitting into the polynomial via regression and minimizing the error. The complexity grows incrementally by trying to add a neuron at a time while the forecasting error is being monitored [19]. When the number of neurons reach a pre-set level, the hidden layer increases. This continues until there is no further performance gain observed, the growth of the polynomial stops and taken as the final equation for the PNN. It is noted that the increment of the network growth is linear.

However, for the case of BFGS-PNN, the expansion of the polynomial is non-linear and heuristic. The optimal state is achieved by unconstrained hill-climbing method guided by Quasi-Network and secant methods. Let an error function be $e(p)$ where $p$ is a vector of real numbers be a vector of network structure information and parameters, i.e. neurons and layers in an ordered set. At the start, when $t=0$, $p_{t=0}$ is initialized by randomly chosen states. Let search direction be $s_i$ at iteration $i$ where time $t=i$. Let $H_i$ be Hessian which is a square matrix of 2$^{nd}$-order partial derivatives of function $e$. where $i$ is the current iteration, $H$ improves becomes a better estimate as the process iterates. $\nabla e(p_i)$ is the gradient of the error function which needs to be minimized at $t=i$ by following the quasi-Newton search pattern using a gradient function similar to Eqn. 2. It seeks to search for the next state of parameters values $p_i+1$ by optimizing $e(p_i+S*s_i)$ where the scalar $S$ must be greater than 0. Eqn. 2 then would have to obey the quasi-Newton condition upon solving the approximation of the Hessian $H_i$, in a way of Eqn. 3.

$$H_i s_i = -\nabla e(p_i) \quad (2)$$
$$H_{i+1}(e_{i+1} - e_i) = \nabla e(p_{i+1}) - \nabla e(p_i) \quad (3)$$

By the secant equation, we let the Hessian matrix take the form as in Eqn. 4.

$$\psi_i = \nabla e(p_{i+1}) - \nabla e(p_i) \text{ and } \varsigma_i = e_{i+1} - e_i, \text{ so } H_{i+1} \text{ satisfies } H_{i+1} * \varsigma_i = \psi_i \quad (4)$$

The updating function for Hessian matrix is defined as Eqn. 5. following the secant method. The equation tries to impose the condition and symmetry such that $\psi_i = H_{i+1}\varsigma_i$. Let $H_{i+1} * \varsigma_i = \psi_i, a = \psi_i \text{ and } b = H_i\varsigma_i$, we can obtain two sub-equations in Eqn. 6. as coefficients to Eqn. 5. Substituting Eqn. 6. Into Eqn. 5, we obtain the updating function for Hessian matrix $H_{i+1}$.

$$H_{i+1} = H_i + \alpha(a \cdot a^\top) + \beta(b \cdot b^\top) \quad (5)$$

$$\alpha = \frac{1}{\psi_i^\top \varsigma_i}, \beta = -\frac{1}{\varsigma_i^\top H_i \varsigma_i} \quad (6)$$

$$H_{i+1} = H_i + \frac{\psi_i \psi_i^\top}{\psi_i^\top \varsigma_i} - \frac{H_i \varsigma_i \varsigma_i^\top H_i^\top}{\varsigma_i^\top H_i \varsigma_i} \quad (6)$$

By applying Sherman-Morrison formula [20] to Eqn. 6., we get Eqn. 7. Which is the inverse of Hessian $H$ matrix. Expanding Eqn. 7 to Eqn. 8 we obtain an equation that can be calculated quickly without needing any buffer space for fast optimization which aims at minimizing $e(*)$.

$$H_{i+1}^{-1} = \left(I - \frac{\varsigma_i \psi_i^\top}{\psi_i^\top \varsigma_i}\right) H_i^{-1} \left(I - \frac{\psi_i \varsigma_i^\top}{\psi_i^\top \varsigma_i}\right) + \frac{\varsigma_i \varsigma_i^\top}{\psi_i^\top \varsigma_i} \quad \text{where } \psi_i^\top H_i^{-1} \psi_i \text{ and } \varsigma_i \psi_i^\top \text{ are scalars} \quad (7)$$

$$H_{i+1}^{-1} = H_i^{-1} + \frac{(\varsigma_i^\top \psi_i + \psi_i^\top H_i^{-1} \psi_i)(\varsigma_i \varsigma_i^\top)}{(\varsigma_i^\top \psi_i)^2} - \frac{H_i^{-1}\psi_i \varsigma_i^\top + \varsigma_i \psi_i^\top H_i^{-1}}{\varsigma_i^\top \psi_i} \quad (8)$$

## 2.2. Fuzzy Rule Induction

By our GROOMS+CMCM methodology, raw data from multiple sources are filtered, condensed, converted into insights of future behaviours in several forms. Traditionally in MC simulation, probability density functions as simulated outcomes are generated, so is sensitivity chart which ranks how each factor in relevance to the predicted outcome. Fuzzy Rule Induction (FRI) plays a role in the methodology by inferring a rule-based model which supplies a series of conditional tests that lead to some consequences based on the same data that were loaded into the MC engine. FRI serves the threefold purpose of easy to use, neural and scalable. Firstly, the decision rules are interpretable by users. They can complement the probability density functions which show a macro view of the situation. FRI helps give another perspective in causality assisting decision makers to investigate the logics of cause-and-effect. Furthermore, different from other decision rule models, FRI allows some fuzzy relaxation in bracketing the upper and lower bounds, thereby a decision can be made based on the min-max values pertaining to each conditional test (attribute in the data). The FRI rules are formatted as branching-logic, which is also known as predicate-logic that preserves the crudest form of knowledge representation. Predicate logic has an IF-TEST[Min-Max]-THEN-VERDICT basic structure and a propensity of how often it exists in the dataset. The number of different groups of FRI rules depend on how many different labels in the predicted class. The second advantage is that the FRI rules are objective and free from human bias as they are derived homogenously from the data. Therefore, they are suitable ingredient for scientifically devising policy and strategy for epidemic control. Thirdly FRI rules can scale up or down not only in quantity, but also in cardinality. A rule can consist of as many tests as the attributes of the data are available. In other words, as a composite MC system, new source of data could be chipped in as per when it becomes necessary or recently available; the attributes of the new data can add on to the conditional tests of the FRI rules.

One drawback about FRI is the lack of indicators for each specific conditional test (or attribute). By the current formulation of FRI, the likelihood of the occurrence of the rule is assigned to the rule as a whole. Little is known about how each conditional test on the attribute is relatively contributing to the outcome that is specified in the rule. In the light of this shortcoming, our proposed methodology suggests that the scores from the sensitivity charts with respect to the relations between the attributes and the outcome, could be used at the rule by simple majority voting.

Rules are generated as a by-product of classification in data mining. The process is through fuzzification of the data ranges and the confidence factors in their effects in classification are taken as indicators. Let a rule be a series of components constraining the attributes $a_{i=1..m}$ (with outcome $y_i = Y$) in the classification model building, so that they can remain valid even though the values are fuzzified. So a rule can be expressed in the predicate rule format such that each $a_i \in \lambda$, where $\lambda \subseteq Y$ is an membership whose labels are mapped to the class labels, a membership takes a range of positive $\mathbb{R}$ where $\lambda = (-\infty, \omega]$. If a rule embraces a conditional test where $(a_i \leqq \omega)$, $\lambda = [\alpha, \infty)$ if it has a test $(ai \geqq \alpha)$. Let $\lambda = [\alpha, \beta]$ when it is comprised of two tests are combined. Replacing the absolute memberships by fuzzy memberships which are defined as fuzzy sets governed by trapezoidal membership function [21]. A fuzzy trapezoidal membership normally would have to be structured by four parameters and expressed as $\lambda^F = (\Theta^{\delta,\downarrow}, \Theta^{\varepsilon,\downarrow}, \Theta^{\varepsilon,\uparrow}, \Theta^{\delta,\uparrow})$. The functional logic is shown in Eqn. 9:

$$\lambda^F(\beta) \stackrel{\text{def}}{=} \begin{cases} 1 & \Theta^{\varepsilon,\downarrow} \leqq \beta \leqq \Theta^{\varepsilon,\uparrow} \\ \frac{\beta - \Theta^{\delta,\downarrow}}{\Theta^{\varepsilon,\downarrow} - \Theta^{\delta,\downarrow}} \Theta^{\delta,\downarrow} < \beta < \Theta^{\varepsilon,\downarrow} \\ \frac{\Theta^{\delta,\uparrow} - \beta}{\Theta^{\delta,\uparrow} - \Theta^{\varepsilon,\uparrow}} \Theta^{\varepsilon,\uparrow} < \beta < \Theta^{\delta,\downarrow} \\ 0 & else \end{cases} \quad (9)$$

where $\Theta^{\varepsilon,\downarrow}$ and $\Theta^{\varepsilon,\uparrow}$ are the lower and upper bounds of the argument $\beta$ which will map to a fuzzy membership of value 1. Similarly, the supports of the lower and upper bounds are denoted by $\Theta^{\varepsilon,\uparrow}$ and $\Theta^{\delta,\uparrow}$.

The fuzzy rules are built on the decision rules which are generated from standard decision tree algorithm, such as direct rule-based classifier equipped with incremental reduced error pruning via greedy search [22]. Given the rule sets generated, the task here is to find the most suitable fuzzy extension for each rule. The task can be seen as replacing the current memberships of the rules by their corresponding fuzzy memberships, which is not too computationally difficult as long as the rule structures and the elements the same. In order to fuzzify a membership, the following formula is applied over the antecedent ($\Omega_i \in \lambda_i$) of rule set while considering the relevant data $X_T^i$; at the same time the instances from the other antecedent ($\Omega_j \in \lambda_j^F$), where $j \neq I$ are to be nullified.

$$X_T^i = \{x | = (x_1 \ldots x_n) \in X_T | \lambda_j^F(x_j) > 0 \; \forall j \neq i\} \subseteq X_T \quad (10)$$

By this approach, the instances $X_T^i$ are divided into two subsets: one subset contains all the positive instances $X_{T+}^i$ and the other subset contains all the negative instances $X_{T-}^i$. After that, a purity measure is used to further separate the two groups into two extremes, positive and negative subgroups, by Eqn. 11.

$$Purity = \frac{\varphi_i}{\varphi_i + (1-\varpi)_i} \quad (11),$$

where positive instances are denoted by $\varphi_i$, and the negative instances are by $\varpi_i$,

$$\varphi_i \stackrel{\text{def}}{=} \sum_{x \in X_{T+}^i} \mu_{a_i}(x), \text{ and } \varpi_i \stackrel{\text{def}}{=} \sum_{x \in X_{T-}^i} \mu_{a_i}(x). \quad (12)$$

When it comes to actual operation, a certainty factor which serves as an indicator about how much the new data instances indeed belong to a subgroup, is needed to quantify the division. Followed by segregating the data into fuzzy rules $\kappa_1^{(j)} \ldots \kappa_r^{(j)}$ by machine learning the relations from their attributes and instance values to some label class $\lambda_j$, further indicator is needed to quantify the strength of each rule. Assume we have a new test instance $x$, Eqn 13 computes the support of the rules of $x$ as follow:

$$S_j(x) \stackrel{\text{def}}{=} \sum_{i=1 \ldots q} \mu_{\kappa_i^{(j)}}(x) \cdot \mathbb{C}(\kappa_i^{(j)}), \quad (13)$$

where $\mathbb{C}(\kappa_i^{(j)})$ is the certainty factor pertaining to the rule $\kappa_i^{(j)}$. The certainty factor $\mathbb{C}$ is expressed in Eqn. 14:

$$\mathbb{C}(\kappa_i^{(j)}) = \frac{2\frac{|X_T^{(j)}|}{|X_T|} + \sum_{x \in X_T^{(j)}} \mu_{\kappa_i^{(j)}}(x)}{2 + \sum_{x \in X_T} \mu_{\kappa_i^{(j)}}(x)}, \quad (14)$$

where $X_T^{(j)}$ denotes the subsets of training instances that are labelled as $\lambda_j$. The result that is predicted by the default classifier to be one of the class labels, is the one that has the greatest value computed from the support function Eqn. 13. At times, some instances $x$ could not be classified into rule or subgroup, that happens when $s_j(x)=0$ for all classes $\lambda_j$, $x$ could be randomly assigned or temporarily placed into a special group. Otherwise, the fuzzy rules are formed, certainty and support indicators are assigned to each one of them. The indicators mean how strong the rules are with respect to the predictive power to the class label possessed by the rules.

## 3. Experiment and Results

For the purpose of validating the proposed GROOMS+CMCM methodology, empirical data proceeding from The Chinese Center for Disease Control and Prevention[‡] (CDCP), an official Chinese government agency in Beijing, China. Since the beginning of the COVID-19 outbreak, CDCP has been releasing the data to the public and updating them daily via mainstream media Tencent and its subsidiary[§]. The data come from mainly two sources: one source is known to be deterministic in nature which is harvested from CDCP in the form of time-series starting from 25 Jan 2020 to 25 Feb 2020. A snapshot of the published data is shown in Figure 2 which are deterministic in nature as historical facts.

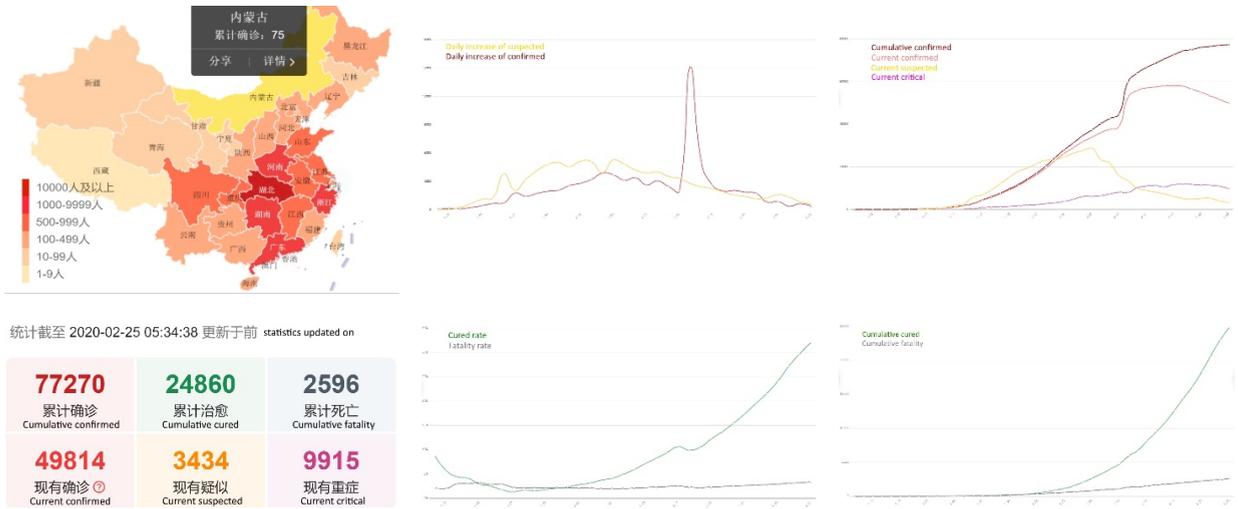

Fig. 2. COVID-19 data disseminated by CDCP with daily updates

### 3.1. Deterministic Input Variables to CMC

The data collected for this experiment are only parts of the total statistics available on the website. The data required for this experiment are the numbers of people in China who have been contracted with the COVID-19 disease in the following ways: suspected of infection by displaying some symptoms, confirmed of infection by medical tests, cumulative confirmed cases, current number of confirmed cases, current number of suspected cases, current number of critically ill, cumulative number of cured cases, cumulative number of deceased cases, recovery (cured) rate % and fatality rate %. This group of time-series are subject to GROOMS for finding the most accurate machine learning technique for obtaining the forecasts as future trends under development. In this case, BFGS-PNN was found to be the winning candidate model, hence applied here for generating future trends for each of the above-mentioned records. The forecasts based on these selected data by BFGS-PNN are shown in Fig 3. The forecasts are in turn used as deterministic input variables to the CMC model. They have relatively lowest errors in RMSE in comparison to other time-series forecasting algorithms as tested in [14]. The rationale is to use the most accurate possible forecasted inputs for achieving the most reliable simulated outcomes from MC simulation at the best effort.

---

[‡] http://www.chinacdc.cn/en/
[§] https://news.qq.com/zt2020/page/feiyan.htm

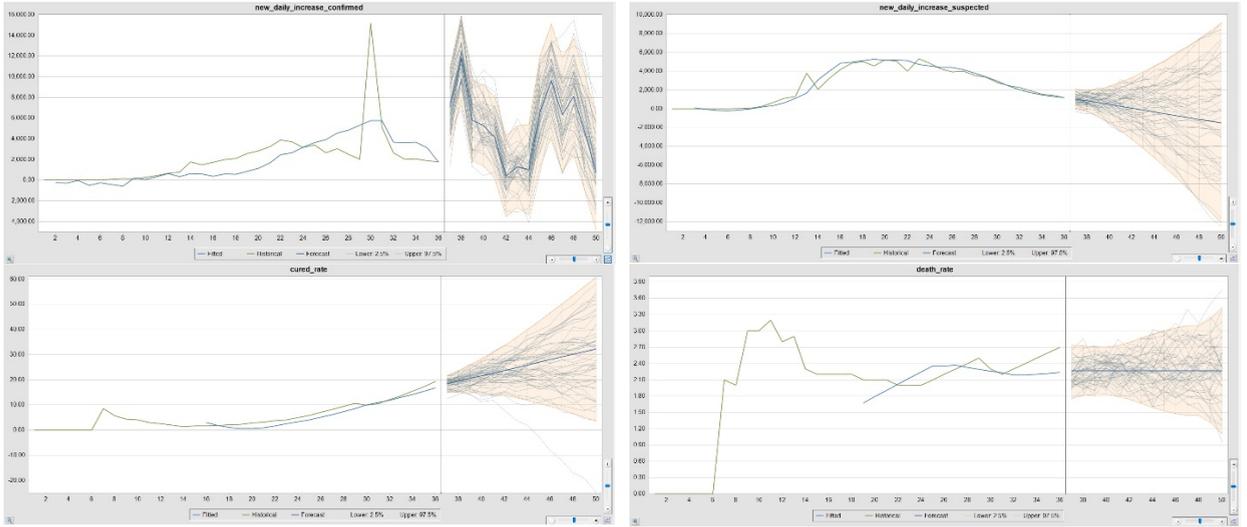

Fig. 3. Forecasts of the main input variables to Monte Carlo simulation using BFGS-PNN

The goal of this Monte Carlo simulation experiment, which is a part of GROOMS+CMCM is to hypothetically estimate the direct cost that is needed as an urgent part of national budget planning to control the COVID-19 epidemic. Direct cost means the cost of medical resources, that includes but not limit to medicine, personnel, facilities, and other medical supplies, directly involved in providing treatments to those patients due to the COVID-19 outbreak. Of course, the grand total cost is much wider and greater than the samples experimented here. The experiment however aims at demonstrating the possibilities and flexibility of embracing both types of deterministic and non-deterministic data inputs by the composite MC methodology.

*3.2. Non-deterministic Input Variables to CMC*

The other group of data that would feed into the CMC are non-deterministic or probabilistic because they would have to bear a high level of uncertainty. They are subject to situation that changes dynamically and there is little control over the outcome. In the case of COVID-19 epidemic control, finding the cure to the virus is a good example. There is best effort put into treatment, but no certainty at all about a cure, let alone knowing when exactly a cure could be developed, tested and proven to be effective for use against the novel virus. There are other probabilistic factors which are used in this experiment as well. Selected main attributes are tabulated in Table 1. We assume a simple equation for estimating the direct cost in fighting COVID-19 using only data of quarantining and isolated medical treatment as follow. Note that the variables shown are abbreviated from the term names. E.g. *d-i-r = days_till_recovery*. The assumptions and hypothesis are derived from past experiences about direct costs involved in quarantine and isolation during the epidemic of SARS in 2003 as published in [24], with reasonable adjustment.

*total_daily_cost* = *cost_for_isolation_till_recovery* + *cost_for_isolation_till_death*
= ((*cured_rate* ÷ (*cured_rate* + *death_rate*) × (*n-d-i-c* × *ppi/day* × *d-f-r*))) +
  ((*death_rate* ÷ (*cured_rate* + *death_rate*) × (*n-d-i-c* × *ppi/day* × *d-i-d*)))

For the non-deterministic variables *ppi/day*, *d-f-r* and *d-t-d*, the following assumptions are derived from [23]. These variables are probabilistic in nature as shown in Table 1. E.g. nobody can actually tell how long an infested patient could be recovered and go home, nor how long the isolation needs to be when the patient is in critical condition. All these are bounded by some probabilities that can be expressed in statistical properties, such as min-max, mean, standard deviation and so on. So some probabilism functions are needed to describe them, and random samples from these probabilities distributions are drawn to run the simulation.

It is assumed that the growth of daily medical cost *ppi/day* follows a normal distribution with daily increase rate. The daily increase rate is estimated from [24] to be rising as days go by, because the Chinese government has been increasingly putting in resources to stop the epidemic using national efforts. The increase is due to the daily increase number of medical staff who are flown to Wuhan from other cities, and the rise of the volume of consumable medical items as well as their inflating costs. The daily cost is anticipated to become increasingly higher as long as the battle against COVID-19 continues at full force. There are other supporting material costs and infrastructure relocation costs such as imposing curfews and economy damages. However, these other costs are not considered for the purpose of demonstrating with a simple and easy-to-understand CMC model.

Normal distribution and uniform distribution are assumed accountable for the increase and probability distributions that describe the lengths of hospital stay. When more information become available, one can consider refining them to Weibull distribution and Rayleigh distribution which can better describe the progress of the epidemic in terms of severity and dual statistical degree of freedom.

*ppi/day* = [Mean: *initial_ppi/day* × (1+*ppi_daily_increase_rate*) ^ #*future_days*, St.Dev=9]
*d-f-r* = [Min: 11, Max: 26]
*d-t-d* = [Mean: 35.9, Std.dev: 6.37]

This is a very simplified approach in guessing the daily cost of the so-called medical expenses, based on only two interventions – quarantine and isolation. Nevertheless, this CMC model though simplified, is serving as an example of how Monte Carlo style of modelling can help generate probabilistic outcomes, and to demonstrate the flexibility and scalability of the modelling system. Theoretically, the CMC system can expand from considering two direct inputs (quarantining and isolation) to 20, or even 200 other direct and indirect inputs to estimate the future behaviour of the epidemic. In practical application, data that are to be considered shall be widely collected, pre-processed, examined for quality check and relevance check (via GROOMS), and then carefully loading into the CMC system for getting the outcomes.

TABLE I: SELECTED INPUT VARIABLES USED IN THE CMC MODEL

| Input variable | Description | Type | Source |
|---|---|---|---|
| new_daily_increase_suspected | The no. of new cases that are suspected to have contracted the COVID-19 virus in a day | Deterministic | www.chinacdc.cn |
| new_daily_increase_confirmed | The no. of new cases that are confirmed to have contracted the COVID-19 virus in a day | Deterministic | www.chinacdc.cn |
| ppq/day | Per person cost of home or secluded camp quarantine per day | Non-deterministic | Estimated by referencing to [23] |
| ppi/day | Per person cost of medical isolation per day | Non-deterministic | Estimated by referencing to [23] |
| cured_rate | The daily rate at which the proportion of patients are cured | Non-deterministic | Estimated by referencing to [23] |
| death_rate | The daily rate at which the proportion of patients died | Non-deterministic | Estimated by referencing to [23] |
| days_for_recovery | The number of days taken for a patient to recover from COVID-19 | Non-deterministic | Estimated by referencing to [23] |
| days_till_death | The number of days taken for a patient to be treated but finally died of COVID-20 | Non-deterministic | Estimated by referencing to [23] |
| cost_for_quarantine | The overall cost for quarantining the necssary number of people per day | Non-deterministic | To be estimated by MC engine |
| cost_for_isolation_till_recovery | The overall cost for isolating and treating the necssary number of people per day until they discharge | Non-deterministic | To be estimated by MC engine |
| cost_for_isolation_till_death | The overall cost for isolating and treating the necssary number of people per day until they died | Non-deterministic | To be estimated by MC engine |
| total_daily_cost | The grand total amount of cost per day for supporting all the quarantine and isolation | Non-deterministic | To be estimated by MC engine |

*3.3. Decision Making under Uncertainty*

Stochastic simulation is well-suited for studying the risk factors of infectious disease outbreaks which always changes in their figures across time and geographical dispersion, thereby posing high level of uncertainty in decision making. Each model forecast by MC simulation is an abstraction of a situation under observation – in our experiment, it is the impact of the dynamics of epidemic development on the direct medical costs against COVID-19. The model forecast depicts the future tendencies in real-life situation rather than statements of future occurrence. The output of MC simulation sheds light in understanding the possibilities of outcomes anticipated.

Being a composite MC model, the ultimate performance of the simulated outcomes would be sensitive to the choice of the machine learning technique that generated the deterministic forecast as input variable to the CMC model. In light of this, a part of our experiment besides showcasing the MC outcomes by the best available technique, is to compare the levels of accuracy (or error) resulted from the wining candidate of GROOMS and a standard (default) approach. The forecasting algorithms in comparison are BFGS+PNN and Linear Regression respectively. The performance criterion is RMSE, which is consistent and unitless $\epsilon_i = log\left(\frac{med(\hat{\theta}_i)}{\theta_i}\right)$ and RMSE=$\left(\sqrt{mean(\epsilon_i^2)}\right)$, as defined in [25]. At the same time, the total costs that are produced manually by explicitly use of spreadsheet using human knowledge are compared vis-à-vis with those of the forecast models by CMCM. The comparative performances are tabulated in Table II. The forecasting period is 14 days. The CMCM model is implemented on Oracle Crystal Ball Release 11.1.2.4.850 (64-bit), running on a i7 CPU @ 2GHz, 16Gb RAM and MS Windows 10 platform. 10,000 trials were set to run the simulation for each model.

TABLE II: COMPARATIVE PERFORMANCE OF VARIOUS FORECAST MODEL APPROACHES

| Forecast model approaches | RMSE | cost_for_quarantine | cost_for_isolation_till_recovery | cost_for_isolation_till_death | total_daily_cost |
|---|---|---|---|---|---|
| Manual forecast using Linear Regression results | -- | 6221267.521 | 70572536.43 | 6074067.024 | 82867870.97 |
| Monte Carlo forecast using Linear Regression results | 127693.55 | 6221267.521 | 101486850.7 | 8017208.093 | 115725326.3 |
| Manual forecast using BFGS+PNN results | -- | 1140569.401 | 44405503.58 | 8135128.297 | 53681201.28 |
| Monte Carlo forecast using BFGS+PNN results | 62077.26 | 1140569.401 | 61556836.31 | 10892459.14 | 73589864.86 |

It is apparent that as seen that from Table II, the RMSE of the Monte Carlo forecasting method using linear regression is more than double that of the method using BFGS+PNN (approximately 128K vs 62K). That is mainly due to the overly estimate of all the deterministic input variables by linear regression. Referring to the first diagram in Figure 3, the variable called *new_daily_increase_confirmed* is non-stationary and it contains an outlier which rose up unusually high near the end. Furthermore, the other correlated variable called *new_daily_increase_suspected*, which is a precedent of the trends of the confirmed cases, also is non-stationary and having an upward trend near the end, though it dips eventually. By using linear regression, the outlier and the upward trends would encourage the predicted curve to continue with the upward trend, linearly, with perhaps a sharp gradient. Consequently, most of the forecast outcomes in the system have been over-forecasted. As such, using linear regression causes unstable stochastic simulation, leading to the more extreme final results, compared to the other methods. This is evident that the *total_daily_cost* has been largely over-forecasted and under-forecasted by manual and MC approaches, in Table II.

On the other hand, when the BFGS+PNN is used, which is able to better recognize non-linear mapping between attributes and prediction class in forecasting, offers more realistic trends which in turn are loaded into the CMCM. As a result, the range between the final *total_daily_cost* results are narrower compared it to its peer linear regression ([LR: 12mil – 83mil] vs [BFGS+PNN: 54mil – 74mil]). The estimated direct medical cost for fighting COVID-19 for a fortnight is estimated to be about 73.6 million USD given the available data using GROOMS+CMCM.

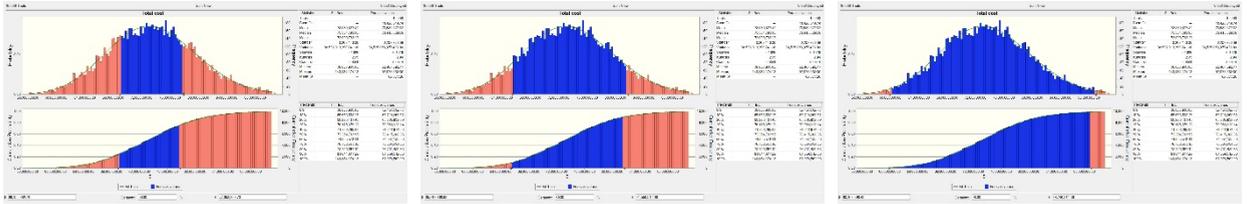

Fig. 4. Probability Distributions as outcomes of CMCM at certainly levels of 50%, 80% and 98% (from far left to right)

According to the results in the form of probability distributions in Figure 4, different options are available for the user to choose from when it comes to estimating the fortnightly budget in fighting this COVID-19 given the existing situation. Each option comes with different price tag, and at different levels of risks. In general, the higher risk that the user thinks it can be tolerated, the lower the budget it will be, and vice-versa. From the simulated possible outcomes in Figure 4, if budget is of constrained, the user can consider bearing the risk (uncertainty of 50%) that a mean of $74mil with [min:$69mil, max:$78mil] is forecasted to be sufficient to fulfil the direct medical cost need. Likewise, if a high certainty is to be assured, for example at 80% the chance that the required budget would be met, it needs about a mean of $79mil with [min:$66mil, max:$82mil]. For a high certainty of 98%, it is forecasted that the budget range will fall within a mean of $90mil and ranging from [min:$60mil, max:$89mil]. As a de-facto practice, some users will take 80% certainty as Pareto Principle (80-20) decision [26] and accept the mean budget at $79mil. $79mil should be a realistic and compromising figure when compared to manual forecast without stochastic simulation, where

$54mil and $84mil budgets would have been forecasted by manual approach by linear regression and neural network respectively.

3.4. Sensitivity Chart and Fuzzy Rules

Sensitivity chart, by its name, displays the extents of how the output of a simulated MC model is affected by changes in some of the input variables. It is useful in risk analysis of a so-called black box model such as the CMCM used in this experiment by opening up the information about how sensitive the simulated outcome is to variations in the input variables. Since the MC output is an opaque function of multiple inputs of composite variables that were blended and simulated in a random fashion over many times, the exact relationship between the input variables and the simulated outcome will not be known except through sensitivity chart. An output of sensitivity chart from our experiment is generated and shown in Figure 5.

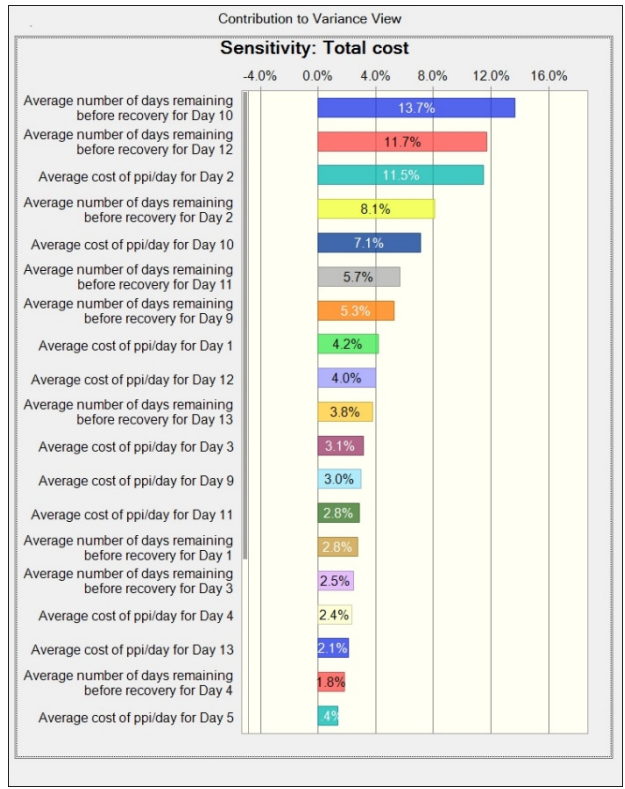

Fig. 5. Sensitivity chart showing the measured sensitivity between each variable and the output

As it can be observed from Figure 5, the top three input variables which are most influential to the predicted output, which is the total medical cost in our experiment are: the average number of days before recovery in Day 10 and Day 12, and average cost per day for isolating a patient in day 2. The first two key variables are about how soon a patient can recover from COVID-19, near the final days. And the third most important variable is the average daily cost for isolating patients at the beginning of the forecasting period. This insight can be interpreted that an early recovery near the final days and reasonably lower medical cost at the beginning would impact the final budget to the greatest extend. Consequently, decision makers could consider that based on the results from the sensitivity analysis, putting in large or maximum efforts in treating isolating patients at the beginning, observe for a period of 10 days or so; if the medical efforts that were invested in the early days take effect, the last several days of recovery will become promising, hence leading to perhaps saving substantially a large portion of medical bill.

Sensitivity chart can be extended to what-if scenario analysis for epidemic key variable selection and modeling [27]. For example, one can modify the quantity of each of the variables, and the effects on the output will be updated instantly. However, this is beyond the scope of this paper though it is worth exploring for it is helpful to fine-tune how the variables should be managed in the effort of maximizing or minimizing the budget and impacts.

Since the effect of a group of independent variables on the predicted output is known and ranked from the chart, it could be used as an alternative to feature ranking or feature engineering in data mining. The sensitivity chart is a by-product generated by the MC after a long trial of repeating runs using different random samples from the input distributions. Figure 6 depicts how the sensitivity chart is related to the processes in the proposed methodology. Effectively the top ranked variables could be used to infer the most influential or relevant attributes from the dataset that loads into an FRI model (described in Section 2.2) for supervised learning. One suggested approach which is fast and easy is to create a correlogram, from there one can do pairwise matching between the most sensitive variables from the non-deterministic data sources to the corresponding attributes from the deterministic dataset. Ranking scores could be mapped over too, by Boyer–Moore majority vote algorithm [28].

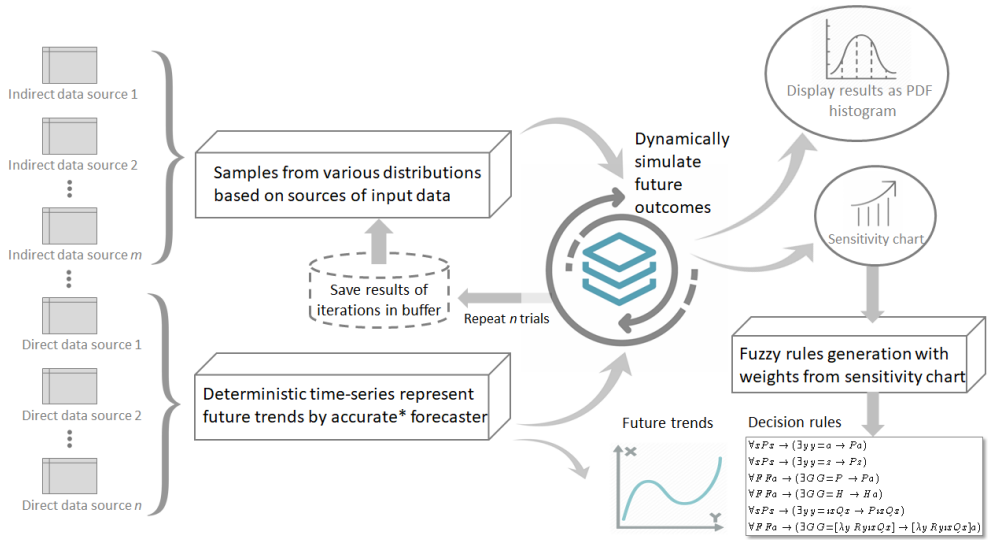

Fig. 6. Sensitivity chart and fuzzy rule induction as outputs from the GROOMS-CMCM methodology (*accuracy by GROOMS)

Some selected fuzzy rules generated by the methodology and filtered by the sensitivity chart correlation mapping are show below. The display threshold is 0.82 which is arbitrary chosen to display only the top six rules where half of them predicts a reflection point for the struggle of controlling the COVID-19 can be attained, the other half indicate otherwise. An FRI model in a nutshell is a classification which predicts an output belonging either one of two classes. In our experiment, we setup a classification model using FRI to predict whether an inflection point of the epidemic could be reached. There is no standard definition of inflection point, though it is generally agreed that is a turning point at which the momentum of accumulation changes from one direction to another or vice-versa. That could be interpreted as either an intersection of two curves of which their trajectories begins to switch. In the context of epidemic control, an inflection point is the moment since when the rate of spreading starts to subside, thereafter the trend of the epidemic is leading to elimination or eradication.

Based on a sliding window of 3 days length, a formula for computing the inflection point based on the three main attributes of the COVID-19 data is given in listed as follow:

Win:
Score  $= w_1 \times$ (Δ down-trend between the past 3 days of *new_daily_increase_confirmed* (n.d.i.c)) +
 $w_2 \times$ (Δ down-trend between the past 3 days of *current_confirmed*) +

$w_3 \times$ (Δ up-trend between the past 3 days of *cured_rate*)

Lose:
Score        = $w_1 \times$ (Δ up-trend between the past 3 days of *new_daily_increase_confirmed* (n.d.i.c)) +
               $w_2 \times$ (Δ up-trend between the past 3 days of *current_confirmed*) +
               $w_3 \times$ (Δ up-trend between the past 3 days of *death_rate*)

where $w_1$=0.1, $w_2$=0.15, and $w_3$=0.25 which can be arbitrarily set by the user. The weights reflect the importance which one considers on how the up or downward trends of confirmed cases and cured vs death rates contribute to reaching the inflection point. A dual curve chart that depicts the inflection point is shown in Figure 7.

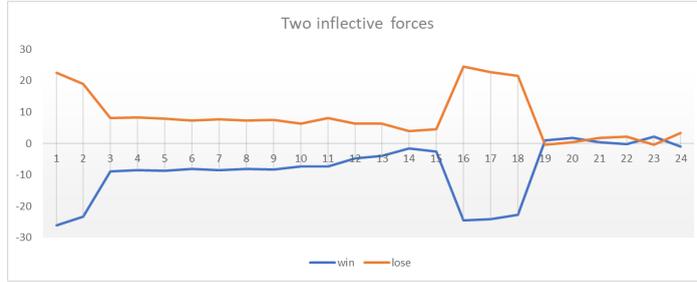

Fig. 7. Curve chart that approximates the search for inflection point during the epidemic

Interestingly, when near the end of the timeline (28/1 – 20/2) that is from point 19[th] onwards, the two curves seem to intervene, as it has been hoping that the winning curve is rise over the losing curve. An inflection point might have reached, but the momentum of the winning is there yet. Further observation on the epidemic development is needed to confirm about the certainty of winning. Nevertheless, the top six rules that are built from classification of inflection point, and processed by feature selection via sensitivity analysis, are shown below. CF stands for confidence factor which indicates how strong the rule is. On the winning side, the rules reveal that when the variables about new confirmed cases fall below certain numbers, a win is scored, contributing towards a reflection point.

Win:
RULE 1:     (*yester3days-ndic* = '(-∞ .. 3581]') → win=1 (CF = 0.96)
RULE 2:     (*ndic* = '(-∞ .. 1874.4]') → win=1 (CF = 0.92)
RULE 3:     (*yester2days-ndic* = '(-∞.. 5027.5]') → win=1 (CF = 0.85)

Lose:
RULE 4:     (*cured_rate* = '(-∞ .. 3.86]') & (*ndic* = '(1874.4 .. 3349.8]') → win=0 (CF = 0.8)
RULE 5:     (*ndic* = '(13677.6 .. ∞)') → win=0 (CF = 0.53)
RULE 6:     (*yester3days-ndic* = '(10706.5 .. ∞)') → win=0 (CF = 0.53)

The strongest rules of the two forces are Rules 1 and 4. Rule 1 shows that to win an epidemic control the down trend over consecutive three days must fall below 3581; on the other hand, the epidemic control may lead to failure, if the cured rate stays less than 3.86% and the new daily increase in confirmed cases remain high between 1874 and 3350 (round up the decimal points).

## 4. Conclusion

Originated from Wuhan, China, the epidemic of novel coronavirus was spreading over many Chinese cities, then over other countries worldwide since December 2019. The Chinese authorities took strict measures to contain the outbreak resolutely by restricting travels suspending business and schools etc. This gives rise to an emergency situation where critical decisions were demanded for, while the virus was novel and very little information was known about the epidemic at the early stage. With incomplete information, limited data on hand, and ever changing on the epidemic development, it is extremely hard for anybody to make a decision using only deterministic approach which foretells precisely the future behaviour of the epidemic. In this paper a composite Monte-Carlo model (CMCM) is proposed to be used in conjunction with GROOMS methodology [23] which finds the best performing deterministic forecasting algorithm. Coupling GROOMS+CMCM together offers the flexibility of embracing both deterministic and non-deterministic input data into the Monte Carlo simulation where random samples are drawn from the distributions of the data from the non-deterministic data sources for reliable outputs. During the early period of disease outbreaks, data are scarce and full of uncertainty. The advantage of CMC is that a range of possible outcomes are generated associated with probabilities. Subsequently sensitivity analysis, what-if analysis and other scenario planning can be done for decision support. As a part of the GROOMS+CMCM methodology, fuzzy rule induction is also proposed, which provides another dimension of insights in the form of decision rules for decision support. A case study of the recent novel coronavirus epidemic (which are also known as Wuhan coronavirus, COVID-19 or 2019-nCoV) is used as an example in demonstrating the efficacy of GROOMS+CMCM. Through the experimentation over the empirical COVID-19 data collected from the Chinese government agency, it was found that the outcomes generated from Monte Carlo simulation are superior to the traditional methods. A collection of soft computing techniques, such as BFGS+PNN, Fuzzy Rule Induction, and other supporting algorithms to GROOMS+CMCM could be able to produce qualitative results for better decision support, when used together with Monte Carlo simulation, than any of deterministic forecasters alone.

# References

1. CDC. 2016. Outbreaks Chronology: Ebola Virus Disease. January 20. Accessed January 22, 2016.
2. The World Bank. 2014. GDP growth (annual %)External. Accessed January 20, 2016.
3. Wuqi Qiu, Cordia Chu, Ayan Mao, and Jing Wu, The Impacts on Health, Society, and Economy of SARS and H7N9 Outbreaks in China: A Case Comparison Study, Journal of Environmental and Public Health, 2018, Volume 2018, Article ID 2710185, 7 pages.
4. William Feuer, Coronavirus: The hit to the global economy will be worse than SARS, NBC Universal, Feb 6 2020
5. Joseph T Wu, Kathy Leung, Gabriel M Leung, Nowcasting and forecasting the potential domestic and international spread of the 2019-nCoV outbreak originating in Wuhan, China: a modelling study, The Lancet, January 31, 2020, DOI:https://doi.org/10.1016/S0140-6736(20)30260-9, pp.1-9
6. Bailey NTJ. 1957 The mathematical theory of epidemics. London, UK: Griffin
7. Andersson H, Britton T. 2000 Stochastic epidemic models and their statistical analysis. Lecture Notes in Statistics, no. 151. Berlin, Germany: Springer
8. Thomas House, Joshua V. Ross and David Sir, How big is an outbreak likely to be? Methods for epidemic final-size calculation, n. Proc R Soc A 469: 20120436, 8 February 2013, https://doi.org/10.1098/rspa.2012.0436
9. Yashima K, Sasaki A (2014) Epidemic Process over the Commute Network in a Metropolitan Area. PLoS ONE 9(6): e98518. https://doi.org/10.1371/journal.pone.0098518
10. Kristin Fitzgerald, Lori Pelletier, and Martin A. Reznek, A Queue-Based Monte Carlo Analysis to Support Decision Making for Implementation of an Emergency Department Fast Track, Journal of Healthcare Engineering, Volume 2017, Article ID 6536523, 8 pages, https://doi.org/10.1155/2017/6536523
11. Bożena Mielczarek and Jacek Zabawa, Monte Carlo Simulation Model to Study the Inequalities in Access to EMS Services, Proceedings 21st European Conference on Modelling and Simulation, EMCS 2007, pp. 1-6
12. Cuesta A., Alvear, D., Abreu O. and Silió D., 2014. Real-time Stochastic Evacuation Models for Decision Support in Actual Emergencies. Fire Safety Science 11: 1063-1076. 10.3801/IAFSS.FSS.11-1063
13. Geoffrey Pettet, Ayan Mukhopadhyay, Mykel J. Kochenderfer, Yevgeniy Vorobeychik, Abhishek Dubey: On Algorithmic Decision Procedures in Emergency Response Systems in Smart and Connected Communities. CoRR abs/2001.07362 (2020)
14. Simon James Fong, Gloria Li, Nilanjan Dey, Rubén González Crespo, Enrique Herrera-Viedma, Finding an Accurate Early Forecasting Model from Small Dataset: A Case of 2019-nCoV Novel Coronavirus Outbreak, International Journal of Interactive Multimedia and Artificial Intelligence, Vol. 6, No. 1, pp.132-140
15. Xia Jiang, Garrick Wallstrom, Gregory F. Cooper, Michael M. Wagner, Bayesian prediction of an epidemic curve, Journal of Biomedical Informatics, Volume 42, Issue 1, February 2009, Pages 90-99
16. Haelterman, Rob (2009). "Analytical study of the Least Squares Quasi-Newton method for interaction problems". PhD Thesis, Ghent University. Retrieved 2014-08-14.
17. Allen, Myron B.; Isaacson, Eli L. (1998). Numerical analysis for applied science. John Wiley & Sons. pp. 188–195. ISBN 978-0-471-55266-6.
18. A. G. Ivakhnenko (1970) "Heuristic Self-Organization in Problems of Engineering Cybernetics". Automatica Vol. 6, pp.207–219

19. A. G. Ivakhnenko, and A. A. Zholnarskiy, (1992) "Estimating the coefficients of polynomials in parametric GMDH algorithms by the improved instrumental variables method", Journal of Automation and Information Sciences c/c of Avtomatika, Vol. 25, no. 3, pp.25-32
20. Sherman, Jack; Morrison, Winifred J. (1949). "Adjustment of an Inverse Matrix Corresponding to Changes in the Elements of a Given Column or a Given Row of the Original Matrix (abstract)". Annals of Mathematical Statistics. 20: 621. doi:10.1214/aoms/1177729959
21. Wing-Kuen Ling, Nonlinear Digital Filters: Analysis and Applications, Academic Press, 2007, ISDN 978-0-12-372536-3
22. Agah, Arvin (2013). Medical Applications of Artificial Intelligence. CRC Press. ISBN 9781439884331. Retrieved 13 August 2017
23. Mubayi A1, Zaleta CK, Martcheva M, Castillo-Chávez C, A cost-based comparison of quarantine strategies for new emerging diseases, Math Biosci Eng. 2010 Jul;7(3):687-717. doi: 10.3934/mbe.2010.7.687
24. Jessica Wang, Ellie Zhu and Taylor Umlauf, How China Built Two Coronavirus Hospitals in Just Over a Week, The Wall Street Journal, Updated Feb. 6, 2020 12:37 pm ET [Last accessed online on 25 Feb 2020]
25. Michael Li, Jonathan Dushoff, and Benjamin M Bolker, Fitting mechanistic epidemic models to data: A comparison of simple Markov chain Monte Carlo approaches, Statistical Methods in Medical Research, 2018, Vol. 27(7) pp.1956–1967
26. Y.-S.Chen, P.P.Chong, M.Y.Tong, Mathematical and computer modelling of the Pareto principle, Volume 19, Issue 9, May 1994, pp.61-80
27. Rimbaud Loup, Bruchou Claude, Dallot Sylvie, Pleydell David R. J., Jacquot Emmanuel, Soubeyrand Samuel and Thébaud Gaël Using sensitivity analysis to identify key factors for the propagation of a plant epidemic5R. Soc. open sci.
28. Boyer, R. S.; Moore, J S. (1991), "MJRTY - A Fast Majority Vote Algorithm", in Boyer, R. S. (ed.), Automated Reasoning: Essays in Honor of Woody Bledsoe, Automated Reasoning Series, Dordrecht, The Netherlands: Kluwer Academic Publishers, pp. 105–117, doi:10.1007/978-94-011-3488-0_5

# Biography

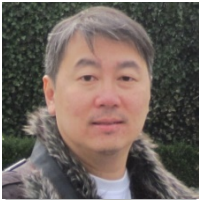

### Simon James Fong

Simon Fong graduated from La Trobe University, Australia, with a 1st Class Honours BEng. Computer Systems degree and a PhD. Computer Science degree in 1993 and 1998 respectively. Simon is now working as an Associate Professor at the Computer and Information Science Department of the University of Macau, as an Adjunct Professor at Faculty of Informatics, Durban University of Technology, South Africa. He is a co-founder of the Data Analytics and Collaborative Computing Research Group in the Faculty of Science and Technology. Prior to his academic career, Simon took up various managerial and technical posts, such as systems engineer, IT consultant and e-commerce director in Australia and Asia. Dr. Fong has published over 450 international conference and peer-reviewed journal papers, mostly in the areas of data mining, data stream mining, big data analytics, meta-heuristics optimization algorithms, and their applications. He serves on the editorial boards of the Journal of Network and Computer Applications of Elsevier, IEEE IT Professional Magazine, and various special issues of SCIE-indexed journals. Simon is also an active researcher with leading positions such as Vice-chair of IEEE Computational Intelligence Society (CIS) Task Force on "Business Intelligence & Knowledge Management", TC Chair of IEEE ComSoc e-Health SIG and Vice-director of International Consortium for Optimization and Modelling in Science and Industry (iCOMSI).

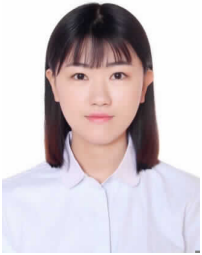

### Gloria Li

Gloria Tengyue Li is currently a PhD student at the University of Macau. She is also the Head of Data Analytics and Collaborative Computing Laboratory, Zhuhai Institute of Advanced Technology, Chinese Academy of Science, Zhuhai, China. Ms Li is leading and managing the laboratory, in R&D as well as technological transfer and incubation. She is an entrepreneur with experiences in innovative I.T. contest, with her award-winning team in the Bank of China Million Dollar Cup competition. Her latest winning work includes the first unmanned supermarket in Macau enabled by the latest sensing technologies, face recognition and e-payment systems. She is also the founder of several Online2Offline dot.com companies in trading and retailing both online and offline. Ms Li is also an active researcher, manager and chief-knowledge-officer in DACC laboratory at the faculty of science and technology, University of Macau.

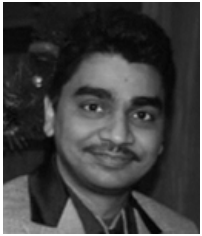

### Nilanjan Dey

Nilanjan Dey, is an Assistant Professor in Department of Information Technology at Techno International New Town (Formerly known as Techno India College of Technology), Kolkata, India. He is a visiting fellow of the University of Reading, UK. He was an honorary Visiting Scientist at Global Biomedical Technologies Inc., CA, USA (2012-2015). He was awarded his PhD. from Jadavpur University in 2015. He is the Editor-in-Chief of International Journal of Ambient Computing and Intelligence, IGI Global. He is the Series Co-Editor of Springer Tracts in Nature-Inspired Computing, Springer Nature, Series Co-Editor of Advances in Ubiquitous Sensing Applications for Healthcare, Elsevier, Series Editor of Computational Intelligence in Engineering Problem Solving and Intelligent Signal processing and data analysis, CRC. He has authored/edited more than 50 books with Springer, Elsevier, Wiley, CRC Press and published more than 300 peer-reviewed research papers. His main research interests include Medical Imaging, Machine learning, Computer Aided Diagnosis, Data Mining etc. He is the Indian Ambassador of International Federation for Information Processing (IFIP) – Young ICT Group.

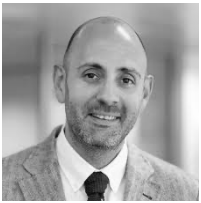

### Rubén González Crespo

Dr. Rubén González Crespo has a PhD in Computer Science Engineering. Currently he is Vice Chancellor of Academic Affairs and Faculty from UNIR and Global Director of Engineering Schools from PROEDUCA Group. He is advisory board member for the Ministry of Education at Colombia and evaluator from the National Agency for Quality Evaluation and Accreditation of Spain (ANECA). He is member from different committees at ISO Organization. Finally, He has published more than 200 papers in indexed journals and congresses.

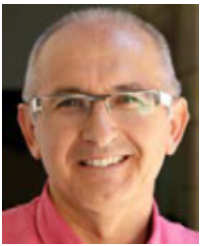

### Enrique Herrera-Viedma

Enrique Herrera-Viedma is Professor in Computer Science and A.I in University of Granada and currently the new Vice-President for Research and Knowledge Transfer. His current research interests include group decision making, consensus models, linguistic modeling, and aggregation of information, information retrieval, bibliometric, digital libraries, web quality evaluation, recommender systems, and social media. In these topics he has published more than 250 papers in ISI journals and coordinated more than 22 research projects. Dr. Herrera-Viedma is Vice-President of Publications of the IEEE SMC Society and an Associate Editor of international journals such as the IEEE Trans. On Syst. Man, and Cyb.: Systems, Knowledge Based Systems, Soft Computing, Fuzzy Optimization and Decision Making, Applied Soft Computing, Journal of Intelligent and Fuzzy Systems, and Information Sciences.